\documentclass{article}
\usepackage{ijcai20}

\usepackage{times}

\usepackage{soul}
\usepackage{url}
\usepackage[hidelinks]{hyperref}
\usepackage[utf8]{inputenc}
\usepackage[small]{caption}
\usepackage{graphicx}
\usepackage{amsmath}
\usepackage{booktabs}
\usepackage{comment}
\title{\LARGE Geo-Spatiotemporal Features and Shape-Based Prior Knowledge\\ for Fine-grained Imbalanced Data Classification}
\author{
$ $ \\
Charles (A.) Kantor*$^{1, 2, 3}$\and
Marta Skreta*$^{1,6}$\and
Brice Rauby$^{2, 3}$\and
Léonard Boussioux$^{2,4}$\and
Emmanuel Jehanno$^{2, 3}$\and
Alexandra Luccioni$^{1, 5}$\and
David Rolnick$^{1,7}$\and
and Hugues Talbot$^{2,3}$\\
\affiliations
$ $ \\
$^1$MILA, Montreal Institute for Learning Algorithms, Montreal, QC, Canada,\\
$^2$Paris-Saclay University, CentraleSupélec Paris (ECP), Greater Paris, France,
$^3$INRIA Paris, France,\\
$^4$Operations Research Center, MIT, Cambridge, MA, USA,
$^5$Université de Montréal, Canada,\\
$^6$University of Toronto, ON, Canada
$^7$McGill University, Montreal, QC, Canada\\
$ $ \\
$ $\texttt{\small ckantor@fas.harvard.edu} \quad \texttt{\small martaskreta@cs.toronto.edu} \quad \texttt{\small leobix@mit.edu}\\
$ $ \\
$ $ \\
$ $ \footnote{Proc. IJCAI 2021, Workshop on AI for Social Good, Harvard University (2021). Copyright by the authors. All rights reserved to authors only. Correspondence to: ckantor (at) fas [dot] harvard [dot] edu}
\textit{\small = equal contribution} \\
$ $ \\
{\small August, 2020} \\

}

\begin{document}

\maketitle 

\begin{abstract}
{\small Fine-grained classification aims at distinguishing between items with similar global perception and patterns, but that differ by minute details. Our primary challenges come from both small inter-class variations and large intra-class variations. In this article, we propose to combine several innovations to improve fine-grained classification within the use-case of wildlife, which is of practical interest for experts. We utilize geo-spatiotemporal data to enrich the picture information and further improve the performance. We also investigate state-of-the-art methods for handling the imbalanced data issue.}
\end{abstract}

\section*{1. \quad Introduction}

While worldwide ecosystems face a mass extinction of species, private and public data related to shifts in species diversity and abundance has substantial taxonomic, spatial, and temporal biases. For illustration purposes, we focus in this work on wildlife monitoring. Insects are vital pollinators, essential to most of our food crops, flowers, and other plants. They represent around 80\% of all animal species and are fundamental to ecosystems. Many insects are important predators of pests in our gardens. They also play a critical role in the recycling of materials, eliminating waste materials, and keeping our soils healthy. A shift in the distribution of species such as bees or butterflies can have severe impacts on human society and environmental equilibria. In \cite{decline2} and \cite{decline}, authors report an alarming decrease in insect populations, as much as 80\% in Europe over the last 30 years. However, this phenomenon is poorly understood, and experts such as entomologists lack large scale data to understand  causes and consequences. There is great potential to efficiently crowdsource and collect at large scale insect abundance data to assess distributional changes and evaluate the impact of climate change and habitat destruction. Identifying an animal to the species or individual level is a challenging task that can rely upon tiny details. Citizen scientists already help collect a large amount of data such as photographic documentation, but accurate identification is a bottleneck. Recent improvements in performance in a wide range of classification tasks with deep learning methods offer new large scale data gathering opportunities. In that context, we develop state-of-the-art computer vision algorithms and propose fine-grained classification innovations: (i) the use of auxiliary data such as geographic location and habitat to further improve performance, and (ii) the exploration of relevant methods for handling imbalanced data, salient for identification. In this paper, we emphasize our work with a citizen science program \cite{ebutterfly}, which maintains a fine-grained dataset of observations of all North American species.

\subsection*{1.1 \quad Related Work}

Fine-grained classification is a category of image classification where the task is to distinguish between subtly rather than grossly different items, like different species of birds or dogs, and unlike giraffes vs. trucks, for example. This task is more complex, requires better annotations, more data and is as of yet not satisfactorily solved~\cite{xie2013hierarchical,chai2013symbiotic}. A key difficulty is to induce the learning architecture to focus on small but important details without relying on overly complex annotations. An interesting recent approach has been to use a deconstruction-reconstruction method to this end~\cite{chen2019destruction}.\\

\paragraph{1.1.1 \quad General and Self-Attention Mechanism}

Attention can be interpreted as a way to focus (or bias) the spatial information of a network onto the areas of an image that seem more relevant to a classification problem~\cite{itti2001computational}. Attention mechanisms have proven very effective in vision and NLP tasks~\cite{vaswani2017attention}. Mechanisms similar to attention but in the channel dimension have been proposed in the form of "squeeze and excitation" networks (SENet)~\cite{hu2018squeeze}. These combined features have been used in fine-grained classification in~\cite{xin2020fine,park2019insect}, particularly, like in our own works \cite{kantorAAAI2020,kantorIAAI2020}, in entomology, zoology and wildlife monitoring. However, in these works, self-attention is used. In our contribution, a prior-shape focus, based on segmentation, is preferred.
\\

\paragraph{1.1.2 \quad Shape-Based Intuition}

Segmentation is a fundamental task in computer vision. Its objective is to find semantically consistent regions that represent objects. A complete review of segmentation methods would require too much space, but given enough data and annotations, deep recurrent CNN architectures such as ResNet~\cite{he2016deep} and recurrent auto-encoders like U-Net~\cite{ronneberger2015u} constitute the current state of the art. Particularly, U-Net and its variants can learn a segmentation task from only a few hundred labeled inputs. The background of macro wildlife sightings is typically full of environmental details like grass or leaves that can mislead the classification model and can introduce bias. Several experiments have revealed that deep networks often pay too much attention to the background instead of the object of interest itself~\cite{eykholt2018robust}. Therefore, automated segmentation to remove or simplify the background, such as in Over-MAP \cite{kantorIAAI2020} is often used as part of uncertainty prediction tool. 

In \cite{kantorAAAI2020}, hierarchical structure of the labels is handled from orders to subspecies. However, even if these hierarchies are typical in the real world, they are difficult to leverage in classification tasks. On the one hand, using the parents-to-children relation seems critical to extract relevant features and to reduce parent-level classification mistakes where the task should be easier. On the other hand, over-penalizing parent-level relationships can cause the classifier to under-perform on leaf classes compared to flat classification. In \cite{kantorAAAI2020}, a loss is designed that enforces the learning of the underlying hierarchy while preserving the flat classification performance.

\paragraph{1.1.3 \quad Additional Features for Visual Classification}

Recent work proposed to integrate complementary information, such as geo-spatiotemporal distribution, to improve classification accuracy \cite{aodha2019presenceonly,chu2019geoaware}. The motivation behind this is that visually similar species may be present in different geographic regions and at different periods of the year, and therefore knowing where and when a picture was taken may be useful information for fine-grained classification. \cite{chu2019geoaware} tested a variety of geo-aware networks and found that incorporating geolocation always showed better performance over the image-only model. \cite{aodha2019presenceonly} developed a geo-spatiotemporal prior that estimates the probability of a species being present based on where and when the image was taken. They showed that incorporating this prior with predictions from an image classifier at test time was able to boost the classification performance of species by 2-12\% depending on the dataset. We use this approach as a motivation to develop a geo-spatiotemporal prior to improve our image classifier.

\subsection*{1.2 \quad Dataset}

In this paper, we illustrate our work with the \textit{City of Montréal}, a collaboration with a large North American's crowdsourcing platform. Citizen scientists recorded sightings by uploading photos with date and time information \cite{ebutterfly}. So far, over 500,000 observations have been submitted across North America, representing over 1000 species as of September 2020. 100,000 of these observations contain images, which were hand-labelled by experts.

Our dataset is organized hierarchically. Each image has mutliple labels (3 of relevance). A label of level 3 belongs to one and only one label of level 2, which also belongs to one and only one family (label of level 1). This distribution of labels enables us to have different levels of complexity for our classification task. Given more than 100,000 labeled images, we anticipate being able to learn the first level (family) label with the best precision, then to provide a slightly less accurate estimate and a slightly worse again estimate of the species (last level considered in this study).

Classes are highly imbalanced in our sample dataset. Indeed a balanced distribution would present linear cumulative distribution functions, while it is not the case here. In previous work~\cite{kantorAAAI2020}, we tackled datasets that were in their majority annotated data by volunteers. In particular a high percentage stemmed from our other collaborators. This data was only partially annotated and some classes were extremely under-represented: it naturally leaded us to consider semi-supervised and few-shot learning. \\

\section*{2. \quad Methods}

\begin{figure}[h]
  \centering
  \includegraphics[scale=0.32]{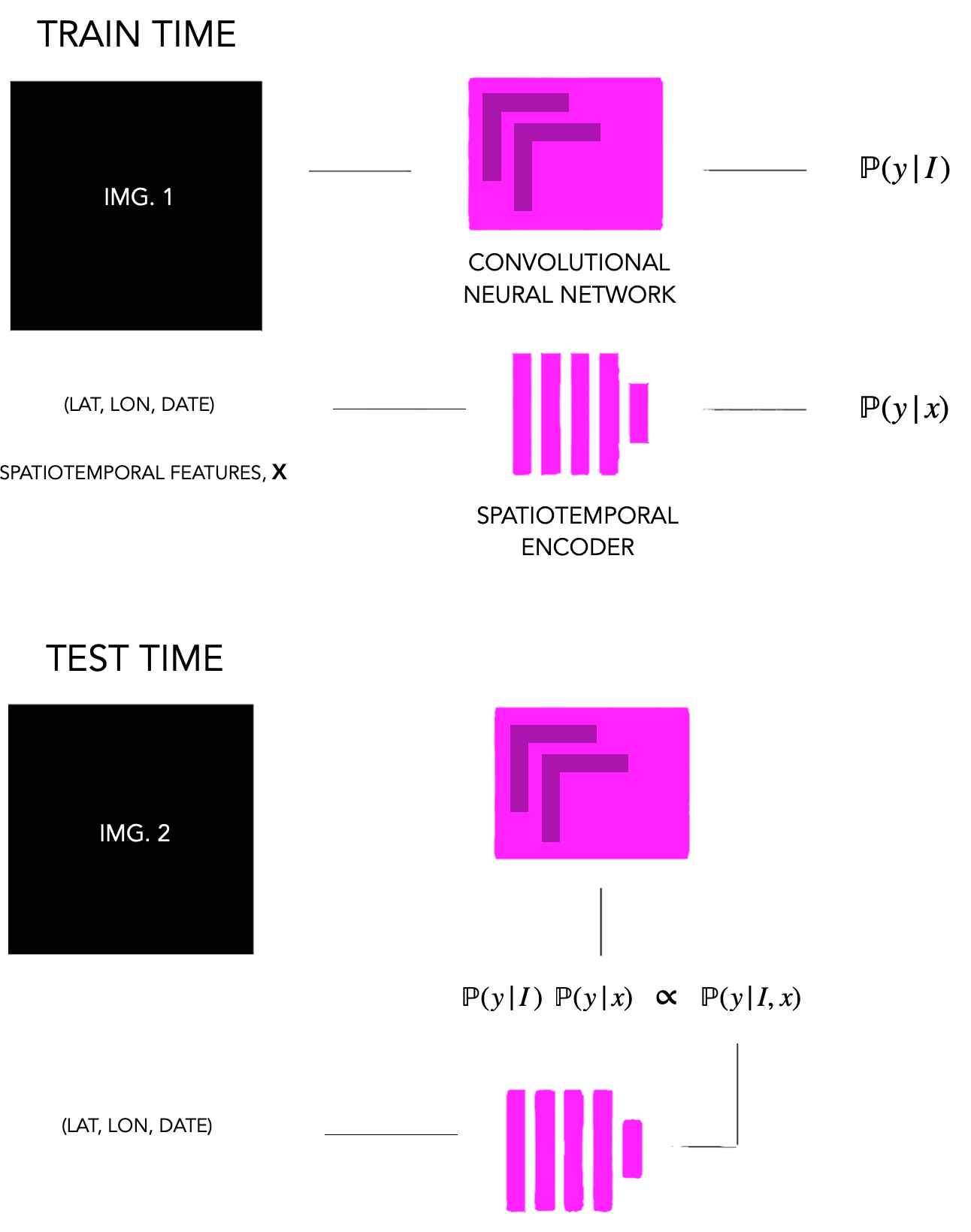}
  \caption{Schematic of incorporating geo-spatiotemporal features. During training, we predict the species from the corresponding image and geo-spatiotemporal data independently. At test time, we use the output from the geo-spatiotemporal model as a Bayesian prior.}
  \label{fig:geomodel_schema}
\end{figure}

Motivated by the approach used by \cite{aodha2019presenceonly}, we also aim at learning a geo-spatiotemporal prior that encodes the presence of a species given the geographic and temporal data associated with the images. This can be helpful to distinguish visually similar species whose geographic ranges do not overlap. We trained two different encoder models (see Figure ~\ref{fig:geomodel_schema}).

\subsection*{2.1 \quad Inclusion of Geo-Spatiotemporal Features}

The first learns the probability of a given species being present in the image with the form 
$P(y|\mathbf{I})$ where $y$ is the class and $I$ is the image. Our image model was built using a CNN residual network architecture \cite{he2016deep} where we trained the final two linear layers from scratch to accommodate the different number of classes gathered. 

The second model is trained to estimate the species from geo-spatiotemporal features: $P(y|\mathbf{x})$ where $x$ is the concatenation of the image's longitude, latitude, and capture date. We transform each of the three features, $x$, using  $[\sin(\pi x), \cos(\pi x)]$ so that longitude and latitude wrap around the Earth and the date wraps around the calendar. Our encoder was the same as in \cite{aodha2019presenceonly}: a series of 9 fully-connected layers with residual links between them. We assume that $\mathbf{x}$ and $\mathbf{I}$ are conditionally independent given the class $\mathbf{y}$. This allows us to use our geo-spatiotemporal model as a Bayesian prior at test time and multiply the probabilities from the image and geo-spatiotemporal models to obtain the final class prediction:

\[P(y|\mathbf{I, x}) \alpha P(y|\mathbf{I})P(y|\mathbf{x})   \]

\[   \]
We computed both the top 1 and top 3 accuracies of this model. We report micro accuracy, which is the total number of correct observations over the total number of observations, and macro accuracy,  which is the average performance of each class (Table \ref{tab:geo_results}). When we incorporated geo-spatiotemporal features into our prior, we saw close to a 2\% improvement in micro accuracy and 6\% improvement in macro accuracy compared to using the images on their own. The latter metric is more appropriate when considering imbalanced datasets such as ours, since it treats majority and minority classes equally. This demonstrates that considering additional features for in-depth classification has the potential to improve the model's final classification performance, especially for underrepresented classes, and calls for further exploration. \\

\begin{table}[h]
  \caption{Model performance on classification model using only images, incorporating coordinates, and incorporating coordinates and time of year. Best accuracies are bolded.}
  \centering
  \begin{tabular}{cccc}%c{1.4cm}c{1.6cm}c{1.9cm}c{2cm}
    \toprule
    Accuracy & Image & Image +  & Image + \\
    & only & (Lat, Lon) & (Lat, Lon, Date) \\
    \midrule
    \centering Top-1, Micro  & 84.56  &  86.16 & \textbf{86.38}\\
    \centering Top-1, Macro  & 59.87 & 64.47 & \textbf{65.31} \\
    \centering Top-3, Micro  & 93.84 & 95.06 & \textbf{95.20}\\
    \centering Top-3, Macro  & 77.53 & \textbf{83.14} & 83.07\\
    \bottomrule
  \end{tabular}
  \label{tab:geo_results}
\end{table}

\subsection*{2.2 \quad Highly Imbalanced Data Approach}

\textit{Since some sightings and thus labels are much rarer than others, as in many real-world problem, dealing with class imbalance is a hard problem that needs to be tackled.}\\

\paragraph*{2.2.1 \quad Generative Models}
A standard setting in semi-supervised learning is to use a model that combines an unsupervised generative component sharing weights with a supervised classifier, for example, using Variational Auto-Encoder as in \cite{VAE} or Generative Adversarial Network \cite{Gan_SSL}. However, the features required for fine-grained classification are different from those required to generate images as suggested from work on Deconstruction and Construction Learning \cite{DCL}. Indeed, features are based on small details, whereas the image generation task has to consider the whole global structure of the image. Some approaches to semi-supervised learning rely on the concept of consistency training: the idea is to make the output labels invariant to the addition of some noise in the input. Consistency training has been originally introduced for data-augmentation in supervised learning. For instance, MixUp \cite{MixUp} generates new images and labels from the convex sum of two images and their corresponding labels. Alternatively, Manifold Mixup \cite{Manifold_Mixup} generalizes it to embeddings within the network. More recently, CutMix \cite{CutMix} has yielded impressive results by substituting a part of the image by the one of another image and performing a linear combination of the corresponding labels based on the proportion of substitution. We applied these methods of consistency training in a semi-supervised configuration as in MixMatch \cite{MixMatch} and Interpolation Consistency Training (ICT) \cite{ICT}. These very similar approaches make use of the MixUp technique using pseudo-labels for unlabelled images.

\paragraph*{2.2.2 \quad Meta and Metric Learning}
Furthermore, a first type of approach which yielded recent improvements is based on the meta-learning paradigm. These methods are derived from the idea of learning the way of updating parameters across different tasks. This is often described as learning how to learn. These approaches make extensive use of episodic training. Episodic training is directly related to transfer learning. The goal is to use a different training set (massively annotated, different from the support set) and to split it into small episodes simulating a few-shot setup. An episode can be seen as a small train and a small test sets. For one episode, the goal is to minimize the generalization error on the test set of the trained model. This model is often simple as in \cite{metalearningwithconvexoptimization} which leverages convex optimization to minimise the generalization of multi-class linear classifier \textit{(multi-class SVM)}. Also, other approaches have been based on graph models to make the best use of the relation between the support set and the query. Indeed, \cite{egnn} introduces an edge-labelling graph to use \textit{intra-cluster} similarity and \textit{inter-cluster} dissimilarity. 

Our Metric Learning approach attempt to learn a distance embedding and then base the classification on the distance between the query embedding and the support set. For example, \cite{prototyping} uses a prototype for each class of the support set and bases the classification on the closest prototype to the query's embedding. Recently,~\cite{local_descriptor} proposed an approach based on local descriptor to replace the image level descriptor used in anterior approaches. Another approach to extract better feature has been proposed in \cite{Cross_attention_network_for_fewshotclassif} and is based on an attention mechanism to focus on features specific to the support set.

\paragraph{2.2.3 \quad Supervised Guidance}
Granted, Few-shot Learning appeared at first sight as a reasonable direction. However, the difference between the few-shot standard configuration and a problem with class imbalance is prohibitive.
Indeed, few-shot learning methods are still designed for balanced datasets. Restricting the training to a few shots when thousands are available (for our most represented levels) appears to be sub-optimal. However, using few-shot learning methods in our setting was designed using a multi-teachers student approach. We train a few-shot algorithm on a truncated version of our dataset and a classical deep learning algorithm on the whole dataset. The student is then optimized to extract features similar to both teachers and reach an optimum individually outperforming the two teachers. Nevertheless, this approach requires training a classifier performing well enough on the under-represented classes, which remains a difficult task. As seen in the previous section, semi-supervised learning methods also require a relatively good classifier when trained in a supervised setting. For this reason, we first designed and train a classifier on our dataset before training it in a semi-supervised manner. 
On the one hand, the recent results yielded by \cite{DCL} seem to discard the use of generative models to solve our problem, as explained in the previous section. On the other hand, consistency training seems to be difficult to apply in our case. Indeed, the fine-grained aspect of the classification task appears as a significant hindrance to design transformation that should preserve the label. For that, we replaced the mix-up transformation by a cut-mix one in the MixMatch semi-supervised learning. However, this approach likely removes discriminative features from one of the two original images. Thus, the fine-grained aspect makes difficult the use of these techniques. In comparison, our teacher-student approaches seem to be more readily applicable in our case and present the advantage of being compatible with few-shot algorithms. To apply such a technique, a teacher that performs decently is necessary. Hence, priority was given to the \textit{supervised approach}.
\\

\paragraph{2.2.4 \quad High-Level Summary}

A common way to overcome \textit{class imbalance} in Machine Learning (ML) is to apply class weights to the model. Also commonly implemented in most ML frameworks, such a \textit{class\_weight} argument is passed to the \textit{fit} function. This argument is a dictionary stating a float value for each class, which corresponds to a penalty parameter to be considered in the computation of the \textit{weighted loss}: a dictionary \{0:1.0, 1:50.0\} forces the model to treat every example of \textit{class 1} as 50 examples of \textit{class 0}.

In the same vein, each class might be sampled differently using a specific strategy based on the number of observations of this class in the dataset. Following an over-sampling strategy, as in the example above where 50 images belong to class 0 and only one to class 1, one might reuse the element from class 1, 50 times. Moreover, with some data augmentation strategies, we get multiple distinct versions from the same image. Nonetheless, the basic information of this one sample likely caused over-fitting. This limitation \textit{(in addition to increased training time)} justifies using different sampling strategies based on the inverse logarithm of the number of occurrences \textit{(presented in the next section)}. Synthetic Minority Over-sampling Technique (SMOTE) introduced in \cite{chawla2002smote} proposes to produce artificial minority samples by interpolating between existing minority samples and their nearest minority neighbors. It is one of the best-known methods to overcome class imbalance. Another approach based on clusters has been proposed in \cite{jo2004class}: minority and majority groups are first clustered using the K-means algorithm, then over-sampling is applied to each cluster separately. This improves both \textit{within-class} imbalance and \textit{between-class} imbalance. Conversely, under-sampling strategies discard data and information about the over-represented class and prevent the model from over-fitting and ignoring some classes altogether \cite{survey_class_imb}.

The results of \cite{10.1145/1273496.1273614} suggest that no sampling method is guaranteed to perform best in all problem domains, and multiple performance metrics should be used when evaluating results. It is also important to keep in mind that these data-level methods can be very time-consuming. Both class weights and sampling strategies can be used together. Ensemble and boosting methods perform well in those situations, such as by iteratively increasing the impact of the minority group by introducing cost items into the AdaBoost algorithm’s weight updates.\\

\paragraph*{2.2.5 \quad Hard Sample Mining and Automated Extraction}
We advanced a performing approach based on Hard Sample Mining, which selects minority samples that are expected to be more informative for each mini-batch, allowing the model to learn more effectively with fewer data. This method presented in \cite{8353718} also uses class rectification loss which is the convex combination of a cross-entropy loss and a triplet loss with a weight depending on the imbalanced property of the class. However, LMLE outperforms CRL in many cases where class imbalance levels are low. We additionally designed a pre-trained U-Net network to generate the segmentation masks and fine-tuned it on a small subset of the dataset. Our approach is possible since the segmentation task is sufficiently similar to the one of segmenting other objects present in a common dataset, thus resulting in a very efficient pre-training (Table 2).

\begin{table}[h]
  \caption{Imbalanced-augmented CNN model performance with and without module for feed-forward visual attention. We provide the average accuracy obtained over 3 different seeds and the standard deviation between parenthesis. Current best accuracies are in bold.}
  \centering
  \begin{small}
  \begin{tabular}{ccccc}
    \toprule
    Accuracy & Augmented & Augmented\\
    & CNN & CNN + Module\\
    \midrule
    Top-1 Acc.  & 79.54 (0.70) & \textbf{80.95 (0.45)}\\
    Top-3 Acc.  & 91.72 (0.49)  & \textbf{93.35 (0.20)}\\
    
    \bottomrule
  \end{tabular}
  \end{small}
  \label{tab:geo_results}
\end{table}

\section*{3. \quad Future work}
We build several novel directions to improve performance further and obtain interesting biological information. After confirming the success of \textit{geo-spatiotemporal features}, we plan to incorporate environmental information about the observed species, such as satellite data which could be used to model habitat. Moreover, each different type of ecosystem can typically host a particular subset of  species. However, this information is not broadly available for butterflies as a preliminary census of the habitat is necessary. Biology empirical knowledge suggests a correlation between the bird and butterfly species found in a given place. We investigate if this hypothesis is correct using bird data as an additional prior to our model. These relationships can be learned and can provide useful information to both entomologists and ornithologists. Another biological use case of our models is to reversely use the masked pictures to focus only on the background. Using a classification algorithm specialized in plants, information can be learned, such as the flowers pollinated or swing plant.

\section*{4. \quad Conclusion}
We have improved the fine-grained classification of taxonomic levels by deep learning, using prior geo-spatiotemporal information. Then, we further investigate state-of-the-art algorithms to handle class imbalance. This current model is now in a deployment on North-American platforms, showing potential for impact.
\\

\bibliography{ijcai20}
\bibliographystyle{named}

\end{document}